\documentclass[runningheads]{llncs}

\usepackage{eccv}

\usepackage{eccvabbrv}

\usepackage{graphicx}
\usepackage{booktabs}

\usepackage[accsupp]{axessibility}  

\usepackage{hyperref}

\usepackage{orcidlink}

\begin{document}

\title{Motion Style Slider: Endpoint-Supervised Continuous Style Control for Human Motion Diffusion} 

\titlerunning{Motion Style Slider}

\author{Chen-Chieh Liao\inst{1,3}\thanks{The work was partially done during an internship at Cygames, Inc.}\orcidlink{0000-0002-9850-2468} \and
Yichen Peng\inst{1}\orcidlink{0000-0002-8544-3905} \and
Yiyi Cai\inst{2}\orcidlink{0009-0008-6078-4536} \and
Y\^ui Ono\inst{3}\orcidlink{0000-0002-8243-7753} \and
Hiroki Hanaoka\inst{3}\orcidlink{0000-0001-8286-5013} \and
Erwin Wu\inst{1}\orcidlink{0000-0002-6723-2864} \and
Hideki Koike\inst{1}\orcidlink{0000-0002-8989-6434} \and
Shuichi Kurabayashi\inst{3}\orcidlink{0000-0001-5967-7727}}

\authorrunning{C.-C.~Liao et al.}

\institute{Institute of Science Tokyo \and
The University of Tokyo \and
Cygames, Inc.}

\maketitle

\begin{abstract}
  Existing human motion diffusion methods provide strong motion generation quality~\cite{tevet2023human}, and recent style transfer models can inject target style cues~\cite{song2024mcmldm,guo2025stylemotif}, but fine-grained \emph{continuous} control of style intensity remains underexplored. In production, style intensity is subjective across artists and directors, so the practical requirement is not a universal absolute unit (e.g., globally correct ``$2\times$''), but a reliable monotonic control axis. We propose \textbf{Motion Style Slider}, a motion-to-motion style transfer framework for endpoint-supervised continuous control. Given a content motion and a style motion, we construct a style direction in a learned motion-style embedding space and condition diffusion generation with a scalar intensity $\alpha$. The training objective combines diffusion denoising with latent intensity regularization to encourage smooth and monotonic style scaling without requiring intermediate-intensity ground-truth motions. Our framework is compatible with pretrained motion diffusion backbones and supports heterogeneous style datasets, including the multi-actor style motion dataset~\cite{kim2025personabooth}. To test out-of-range usability, we additionally introduce a small real-capture over-reaction extension and evaluate large-$\alpha$ behavior against these unseen targets. Experiments measure controllability, interpolation/extrapolation behavior, content preservation, and motion realism, with ablations on direction construction and loss design.
\end{abstract}

\section{Introduction}
\label{sec:intro}

\begin{figure}[t]
  \centering
  \includegraphics[width=\linewidth]{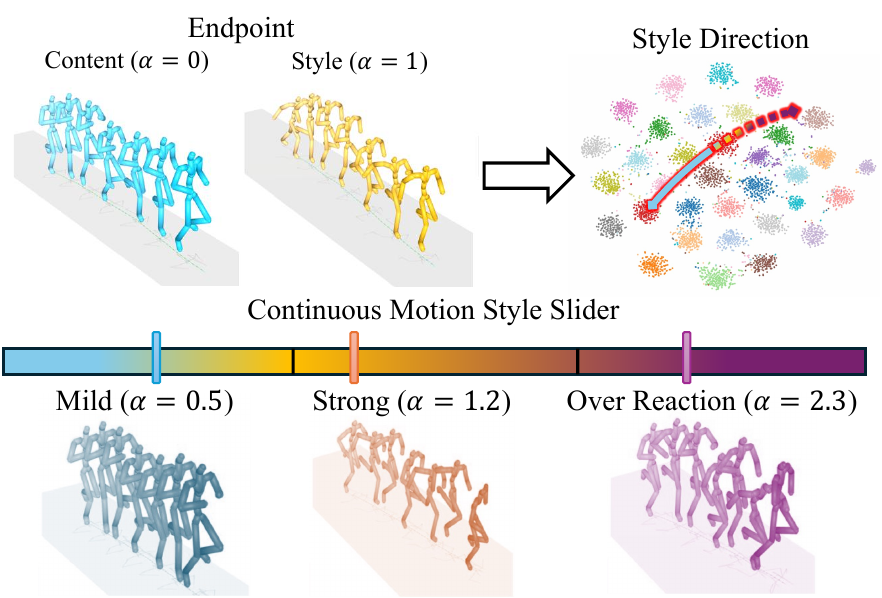}
  \caption{\textbf{Motion Style Slider overview.} Given a content motion and a target style endpoint, our model produces a continuous family of motions controlled by intensity $\alpha$, from neutral-style behavior ($\alpha=0$) to stylized behavior ($\alpha=1$) and extrapolated over-reaction ($\alpha>1$).}
  \label{fig:intro_overview}
\end{figure}

Generating expressive human motion with controllable style is a central goal for animation, virtual characters, and interactive content creation. Recent motion diffusion models have substantially improved realism and diversity~\cite{tevet2023human,zhang2022motiondiffuse,guo2022tm2t,guo2022generate}, while motion style transfer methods have shown strong adaptation from reference examples~\cite{song2024mcmldm,guo2025stylemotif,yang2025contrastive}. However, most existing pipelines still treat style control as a discrete problem (e.g., \textit{happy}, \textit{sad}) or as weak latent interpolation without explicit intensity supervision. As a result, fine-grained style strength control remains unreliable, and extrapolation beyond seen style strength is rarely validated with real motion targets.

In real game-animation workflows, this appears as iterative direction from directors and animation leads: ``more like this'', ``a bit less'', ``push it further''. Importantly, style intensity is not absolute across users: one person's ``2$\times$'' may be another's ``1.5$\times$''. This makes relative controllability the key requirement: as the slider increases, style should change in a predictable order, with smooth transitions and moderate extrapolation headroom beyond observed endpoints.

To address this gap, this paper focuses on \textbf{continuous style intensity control} for human motion diffusion. We seek a practical setting where only endpoint supervision is available: a neutral-content motion and a fully stylized motion with the same underlying content. Intermediate-intensity motions are generally unavailable in existing datasets, yet these intermediate states are exactly what users need for controllable generation. This creates a key learning challenge: how to produce smooth, monotonic, and extrapolatable style variation without intermediate ground-truth.

As shown in Figure~\ref{fig:intro_overview}, our central idea is to model style change as a \textbf{direction} in a learned style embedding space and to use an explicit scalar intensity variable $\alpha$ for conditioning. Concretely, we compute a style direction from endpoint motions and construct a continuous conditioning code by moving along this direction. The diffusion model receives explicit content-motion conditioning together with style-intensity conditioning, and is trained with denoising loss plus latent intensity regularization. This design makes style intensity control explicit and testable while preserving input content structure.

In this work, we implement this idea on top of an off-the-shelf motion diffusion framework, with style representations derived from motion-language alignment models. Our training and evaluation cover multiple style datasets, enabling analysis across heterogeneous style taxonomies and capture conditions. To directly test out-of-range behavior, we additionally collect a small over-reaction extension with stronger-than-endpoint captures, used as unseen high-strength targets.

Compared with prior motion style transfer pipelines that emphasize one-shot transfer quality, our goal is reliable relative control. We therefore center evaluation on: (i) style monotonicity with respect to $\alpha$, (ii) interpolation and modest out-of-range behavior, (iii) content preservation, and (iv) motion realism.

In summary, our main contributions are:
\begin{itemize}
\item We formulate diffusion-based motion-to-motion style transfer as an \textbf{endpoint-supervised continuous control} problem, where style intensity is explicitly parameterized by a scalar slider $\alpha$ and interpreted as a relative control coordinate.
\item We propose a style-direction conditioning and latent regularization scheme that enables smooth and monotonic style scaling without requiring intermediate-intensity ground-truth motions.
\item We provide a multi-dataset evaluation protocol, including a new real-capture over-reaction split for out-of-range analysis, and report ablations that isolate the effects of direction design and loss terms.
\end{itemize}

\section{Related Work}
\label{sec:related}

\subsection{Text-to-Motion Generation}

Text-to-motion (T2M) generation has progressed from sequence modeling and motion tokenization to diffusion-based synthesis.
Early works introduce motion tokenization and reciprocal text–motion generation frameworks such as TM2T and T2M~\cite{guo2022tm2t,guo2022generate}.
Building on discrete motion tokens, subsequent works explore language–motion large models including MotionGPT, T2M-GPT, MotionGPT-2, and unified large motion models~\cite{jiang2024motiongpt,zhang2023t2mgptgeneratinghumanmotion,wang2024motiongpt2,zhang2024largemotionmodel}, which treat human motion as a language-like sequence for autoregressive generation.
Other extensions further explore masked motion modeling, retrieval augmentation, efficient sequence modeling, keyframe control, or expressive whole-body generation~\cite{guo2024momask,zhang2023remodiffuse,zhang2024motionmamba,geng2024keymotion,liu2024t2mx,cai2026flooddiffusiontailoreddiffusionforcing}.

More recently, diffusion-based approaches significantly improved motion realism and diversity.
Representative methods include MotionDiffuse~\cite{zhang2022motiondiffuse}, MDM~\cite{tevet2023human}, retrieval-augmented and prior-based variants~\cite{zhang2023remodiffuse,shafir2024human}, and latent diffusion frameworks such as MLD~\cite{song2024mcmldm}.
These models learn strong motion priors from large text–motion datasets and enable high-quality motion synthesis conditioned on textual descriptions; recent controllable variants further incorporate spatial constraints or real-time control~\cite{karunratanakul2023gmd,xie2023omnicontrol,dai2024motionlcm}.

Despite these advances, most T2M systems primarily control action semantics (e.g., ``walk'' or ``jump'') rather than the intensity or strength of style.
Consequently, they provide limited mechanisms for calibrated stylistic variation.

\subsection{Motion Style Transfer}

Motion style transfer (MST) aims to modify the stylistic characteristics of a motion sequence while preserving its underlying content. 
Styles may correspond to emotional expressions, performer-specific traits, or character-driven motion patterns, enabling expressive virtual characters and animation systems.

Style transfer was first popularized in the image domain~\cite{Gatys2016,Johnson2016,Huang2017,Choi2020,Diederik2018}, where neural networks disentangle content and style representations and recombine them to synthesize stylized outputs. 
These ideas later inspired motion-based approaches~\cite{Holden2015,Holden2016,Du2019,Aberman2020,Park2021,Yu-Hui2021}. 
Early methods relied on optimization-based frameworks that iteratively adjusted motion features to match target style statistics~\cite{Holden2016}, while subsequent learning-based models improved efficiency and scalability~\cite{Holden2016,Du2019}. 

Later work explored more flexible stylization strategies, including label-free or weakly supervised style learning~\cite{Yu-Hui2021,jang2022motion}, cross-content style transfer~\cite{kim2024most, liao2023}, spatial–temporal modeling and stochastic style generation~\cite{Park2021}, kinematic constraints for physically plausible motion~\cite{Huaijun2021}, and real-time stylization frameworks for interactive applications~\cite{mason2022realtime}.

More recently, diffusion-based generative models have significantly improved stylization fidelity and motion realism. 
Diffusion-based MST methods~\cite{hu2024diffusionbasedhumanmotionstyle,zhong2024smoodi,qian2024smcd} leverage strong generative priors to synthesize high-quality stylized motions, while domain-specific approaches such as dance stylization further extend these ideas to specialized motion domains~\cite{sawdayee2025dance}. 
Latent diffusion frameworks further enable flexible motion–style combinations using multi-condition representations, including MCM-LDM~\cite{song2024mcmldm} and StyleMotif~\cite{guo2025stylemotif}.

Despite these advances, most existing methods treat style as a categorical condition or perform transfer toward a single target style instance. 
As a result, stylization strength is typically evaluated at a fixed operating point, and continuous or calibrated control of style intensity remains largely unexplored.

\subsection{Motion Representation Learning and Style Conditioning}

Learning structured motion representations is crucial for controllable motion generation and stylization.
Recent work has explored cross-modal representation learning between motion and language to obtain semantically meaningful embedding spaces.

MotionCLIP~\cite{tevet2022motionclip} aligns motion representations with CLIP features to enable text-driven motion editing and retrieval.
Similarly, TMR~\cite{petrovich23tmr}, building on text--motion representation learning such as TEMOS~\cite{petrovich2022temos}, learns retrieval-oriented text–motion embeddings that capture semantic relationships between motions and language descriptions.
Such embedding spaces provide a natural basis for defining semantic directions or conditioning signals for controllable motion generation.
We use a TMR-based motion encoder because it provides a frozen, motion-level semantic space with stable retrieval behavior across different actions and styles. 
Our contribution is not to introduce a new representation model; instead, we use this representation to define endpoint style directions, train a controllable diffusion adaptor, and evaluate whether generated motions move predictably along the intended direction.

Another line of work studies actor-specific or persona-oriented motion generation.
Persona-based models aim to capture performer-specific motion characteristics and enable personalized motion synthesis~\cite{kim2025personabooth}.
Datasets such as PerMo highlight the importance of modeling stylistic variation across actors and performing styles. Our work is complementary to these approaches.
Rather than focusing solely on personalized motion generation, we study the problem of \textbf{continuous style intensity scaling}.
To evaluate this property, we conduct experiments across multiple style datasets~\cite{kim2025personabooth,kobayashi2023motion,xia2015dataset} and further introduce a real-capture ``over-reaction'' split to evaluate extrapolation beyond observed style endpoints.
This setup emphasizes that controllable stylization should be assessed as a continuous and extrapolatable property rather than a single-point style transfer result.

\section{Method}
\label{sec:method}

\begin{figure}[t]
  \centering
  \includegraphics[width=\linewidth]{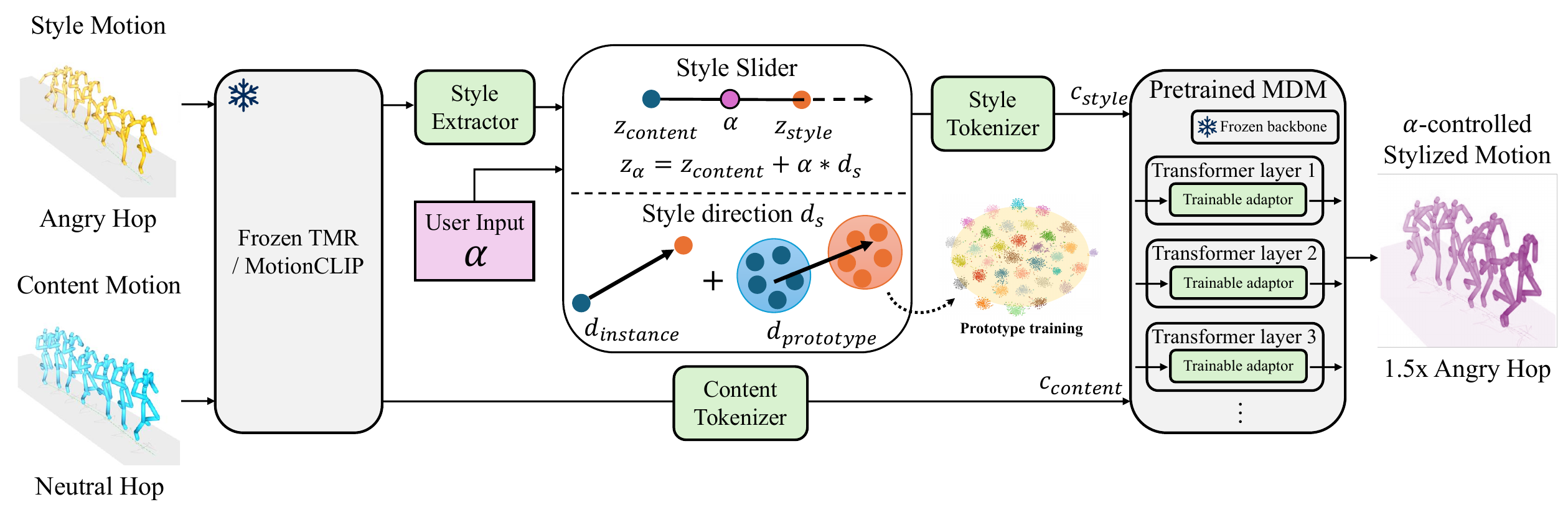}
  \caption{\textbf{Method pipeline.} Given endpoint motions, we compute a style direction in latent space, sample an intensity value, build a style condition token, and train a diffusion denoiser with endpoint-aware supervision and latent intensity regularization.}
  \label{fig:method_pipeline}
\end{figure}

\subsection{Problem Formulation}

As illustrated in Figure~\ref{fig:method_pipeline}, we study motion style control under an \emph{endpoint-supervised} setting. 
Let $m_c \in \mathbb{R}^{T\times D}$ denote a neutral or content motion, and $m_s \in \mathbb{R}^{T\times D}$ denote its stylized endpoint corresponding to the same content instance. 
Training data provides only such endpoint pairs $(m_c,m_s)$, while intermediate-intensity motions are typically unavailable.

Our goal is to learn a model that generates motion $\hat{m}_\alpha$ for a continuous intensity parameter $\alpha \ge 0$ such that
\begin{equation}
\begin{aligned}
\hat{m}_{\alpha=0} &\approx m_c, \\
\hat{m}_{\alpha=1} &\approx m_s,
\end{aligned}
\end{equation}

where $\hat{m}_{\alpha} \text{ changes style smoothly and monotonically with } \alpha$.
Importantly, $\alpha$ represents a \emph{relative control coordinate} rather than an absolute style scale. 
This design reflects practical animation workflows, where style strength is adjusted incrementally rather than specified by a fixed universal unit. 
To implement this control mechanism, we build on a pretrained MDM diffusion model~\cite{tevet2023human} and introduce conditioning and training strategies that enable continuous style modulation while preserving the underlying motion content.

\subsection{Style Direction in Latent Space}

To represent style variation in a structured manner, we operate in a learned motion embedding space. 
Let $E_s(\cdot)$ denote a frozen style encoder (a TMR-based head~\cite{petrovich23tmr} in our implementation) that maps a motion sequence to a normalized embedding:

\begin{equation}
z = E_s(m), \qquad z \in \mathbb{R}^{d}, \qquad \|z\|_2 = 1 .
\end{equation}

Working in this latent space allows style manipulation to occur along semantically meaningful directions rather than directly modifying raw joint coordinates. 
The normalization stabilizes direction scale across different motions, so the scalar $\alpha$ controls movement along the style direction rather than being dominated by embedding magnitude.

For each training pair, we compute the embeddings of the content and stylized endpoints:

\begin{equation}
z_c = E_s(m_c), \qquad z_s = E_s(m_s).
\end{equation}

These two points define a local style axis associated with the pair. 
We estimate a style direction $d_s$ using one of three strategies:

\begin{equation}
d_{\text{instance}} = z_s - z_c, 
\qquad
d_{\text{prototype}} = \mu_s - \mu_c,
\end{equation}

\begin{equation}
d_s =
\begin{cases}
d_{\text{instance}}, & \text{instance mode},\\
d_{\text{prototype}}, & \text{prototype mode},\\
(1-\beta)d_{\text{prototype}} + \beta d_{\text{instance}}, & \text{hybrid mode}.
\end{cases}
\end{equation}

Here $\mu_s$ and $\mu_c$ denote style and content/neutral prototypes computed from the training data, and $\beta \in [0,1]$ balances global and instance-specific style directions.

\subsection{Condition Construction}

To preserve motion content during stylization, we explicitly provide a content-motion condition to the diffusion model. 
Let $E_c(\cdot)$ denote a frozen motion encoder that extracts a content representation:

\begin{equation}
h_c = E_c(m_c), \qquad h_c \in \mathbb{R}^{d_c}.
\end{equation}

In our implementation, $E_c$ corresponds to the TMR motion encoder with masked mean pooling over motion tokens (excluding the CLS token). 
We project this representation into a diffusion conditioning token:

\begin{equation}
c_{\text{content}} = P_c(h_c) \in \mathbb{R}^{1 \times d},
\end{equation}

where $P_c$ is a learnable projection module. 
The encoder $E_c$ remains frozen to maintain stable semantics, while $P_c$ and lightweight MDM adaptor layers are trained to inject content and style conditions into the pretrained denoiser.

Given an intensity value $\alpha$, we construct the style condition by moving along the style direction:

\begin{equation}
z_{\alpha} = z_c + \alpha d_s.
\end{equation}

This equation implements the \emph{style slider}: increasing $\alpha$ moves the embedding along the style axis, continuously strengthening the stylization relative to the content anchor $z_c$. 
In implementation variants, we optionally normalize the condition vector or direction for numerical stability.
The resulting vector is projected into a style conditioning token $c_{\text{style}}$. 
Both content and style tokens are then provided to the MDM denoiser:

\begin{equation}
\hat{x}^{0}_{\theta}(x_t, t \mid c_{\text{content}}, c_{\text{style}}, c_{\text{text}}).
\end{equation}

Text conditioning $c_{\text{text}}$ is optional and disabled in our default configuration.

During training, we sample the intensity value $\alpha$ using an endpoint-aware scheme:

\begin{equation}
\alpha =
\begin{cases}
0, & \text{with probability } p_0, \\
1, & \text{with probability } p_1, \\
u,\; u \sim \mathcal{U}(0,\alpha_{\max}), & \text{otherwise}.
\end{cases}
\end{equation}

This sampling strategy increases the frequency of endpoint supervision while still exposing the model to intermediate intensity values, improving the reliability of the slider behavior.

\subsection{Training Objective}

For each sampled $\alpha$, we define the diffusion target motion as

\[
m_\alpha^\star =
\begin{cases}
m_c, & \alpha = 0,\\
m_s, & \alpha > 0.
\end{cases}
\]

The standard diffusion denoising objective is then

\begin{equation}
\mathcal{L}_{\text{diff}}
=
\mathbb{E}_{m^\star_\alpha,t,\epsilon}
\left[
\left\|
x^0 - \hat{x}^{0}_{\theta}(x_t,t,c_{\text{content}},c_{\text{style}},c_{\text{text}})
\right\|_2^2
\right].
\end{equation}

To maintain strong endpoint fidelity, we apply different weights to endpoint and mid-range samples: endpoints use weight 1, while intermediate samples are down-weighted by $w_{\text{mid}}$.
Because no real midpoint motions are available, intermediate-$\alpha$ samples are not treated as hard motion-space midpoint targets; instead, the diffusion loss preserves endpoint realism while the latent losses below shape the generated style trajectory.

While the diffusion loss ensures motion realism and endpoint reconstruction, it does not explicitly enforce consistent behavior along the style axis. 
We therefore introduce two additional regularization terms.

First, we encourage linear scaling of style strength with respect to $\alpha$. 
Using the endpoint direction $\Delta = z_s - z_c$, we define a style projection function

\begin{equation}
g(m) =
\left\langle
E_s(m) - z_c,
\frac{\Delta}{\|\Delta\|_2}
\right\rangle.
\end{equation}

For generated motion $\hat{m}_\alpha$, the linearity constraint becomes

\begin{equation}
\mathcal{L}_{\text{lin}} =
\left(g(\hat{m}_\alpha) - \alpha g(m_s)\right)^2.
\end{equation}

Second, we enforce monotonicity along the style axis. 
Given two sampled intensities $\alpha_2 > \alpha$, we penalize violations of increasing style strength:

\begin{equation}
\mathcal{L}_{\text{mono}} =
\max\!\left(0,\gamma - \left(g(\hat{m}_{\alpha_2}) - g(\hat{m}_{\alpha})\right)\right).
\end{equation}

Here $\gamma \ge 0$ is a margin encouraging separation between different intensity levels. 
This loss discourages cases where larger $\alpha$ unexpectedly produces weaker stylization.

The final objective combines these components:

\begin{equation}
\mathcal{L}
=
\mathcal{L}_{\text{diff}}
+
\lambda_{\text{lin}}\mathcal{L}_{\text{lin}}
+
\lambda_{\text{mono}}\mathcal{L}_{\text{mono}} .
\end{equation}

In practice, $\mathcal{L}_{\text{lin}}$ and $\mathcal{L}_{\text{mono}}$ are applied to non-endpoint sampled $\alpha$ values, including extrapolation samples when $\alpha>1$. 
Together, $\mathcal{L}_{\text{diff}}$ maintains motion realism, while the additional regularizers shape the controllability of the style slider.

\subsection{Training and Inference}

Training uses endpoint pairs with sampled $\alpha$ values and chosen direction modes (instance, prototype, or hybrid). 
Model parameters are optimized with AdamW starting from pretrained MDM weights. 
In our main configuration, we use adaptor-only finetuning with motion-conditioned content input and the text branch disabled.

At inference time, given a pair $(m_c,m_s)$ and a user-selected intensity $\alpha$, we compute the corresponding style direction, construct the conditioning token, and sample motions from the diffusion model. 
The output duration follows the content motion mask; sequences are padded or truncated only for batching, and evaluation is performed on the valid frames.
By sweeping $\alpha$ from 0 to values above 1, the system produces both interpolation and extrapolation trajectories along the learned style axis.

\section{Experiments}
\label{sec:exp}

\subsection{Datasets}
We evaluate on three benchmark style-motion datasets: PerMo~\cite{kim2025personabooth}, Bandai-Namco~\cite{kobayashi2023motion}, and Xia~\cite{xia2015dataset} (Table~\ref{tab:dataset_summary}). All data are converted to the same feature representation as proposed in \cite{guo2022generate}, and endpoint pairs $(m_c,m_s)$ are built where $m_c$ is neutral/content and $m_s$ is stylized.

For extrapolation, we build a game-industry-oriented motion set captured with professional actors, following a protocol and style-definition process similar to Bandai-Namco. During capture, actors are explicitly instructed to perform over-reactions to simulate intensified style motions.
We use this set as matched triplets $(m_c,m_s,m_{s_2})$, where $m_{s_2}$ is a stronger real capture of the same content-style identity. These clips are test-only and excluded from training. We retarget raw captures to the same skeleton/feature space and keep only endpoint-consistent valid triplets after quality-control filtering.

Throughout the experiment, we use canonical train/val/test splits and evaluate on test pairs only.

\begin{table}[t]
  \centering
  \caption{\textbf{Dataset summary.} Statistics from current canonical splits and our over-reaction extrapolation pairs used in this paper.}
  \label{tab:dataset_summary}
  \begin{tabular}{lcccl}
    \toprule
    Dataset & \#Styles & \#Contents & \#Clips & Protocol Use \\
    \midrule
    PerMo & 33 & 10 & 13,220 & main + ablation \\
    Bandai-Namco & 14 & 1 & 175 & main \\
    Xia & 7 & 5 & 572 & main \\
    Ours (over-reaction) & 2 & 5 & 188 & extrapolation-only \\
    \bottomrule
  \end{tabular}
\end{table}

\subsection{Implementation Details}
All methods are evaluated with the same canonical pair protocol and the same intensity set $\alpha \in \{0,0.5,1.0,1.5,2.0\}$ (or ablation-specific subsets). Unified metrics are computed with a shared evaluator using a frozen TMR style encoder head as the feature space.

For our method, default main settings are: instance direction ($d_s=d_{\text{instance}}$), adaptor-only finetuning, pooled motion-content conditioning, and text branch disabled. Prototype construction details are provided in the supplementary material.

Baselines are evaluated on the same pair splits and alphas, using their own trained checkpoints and the same post-hoc evaluator. This keeps the metric extractor and prototype construction identical across methods.

\subsection{Baselines and Variants}
We compare against DeepMotionEditing (Aberman et al.)~\cite{Aberman2020} and MCM-LDM~\cite{song2024mcmldm}. Because these methods expose different controls, we harmonize inputs to endpoint pairs and evaluate with the same pair list, alphas, and feature-space metrics.
In particular, MCM-LDM is a strong latent-diffusion style-transfer baseline, but it does not explicitly optimize an endpoint-anchored style direction or a scalar intensity-control objective. 
This distinction is central to our evaluation: we compare not only endpoint transfer quality, but also whether a method responds predictably as $\alpha$ changes.
Additional evaluation on the PerMo dataset with MoST~\cite{kim2024most}, SMooDi~\cite{zhong2024smoodi}, MotionCLIP~\cite{tevet2022motionclip}, and the HumanML3D T2M evaluator~\cite{guo2022generate} are provided in the supplementary material.

For ablation analysis, we evaluate:
\begin{itemize}
\item Direction mode: prototype/instance/hybrid,
\item Objective components: with/without $\mathcal{L}_{\text{lin}}$ and $\mathcal{L}_{\text{mono}}$.
\end{itemize}

\subsection{Metrics}
We evaluate:
\begin{itemize}
\item \textbf{Content preservation}: Content Recognition Accuracy (CRA), using nearest prototype in content embedding space.
\item \textbf{Style fidelity}: Style Recognition Accuracy (SRA), using nearest prototype in style embedding space.
\item \textbf{Motion realism}: Fr\'echet Motion Distance (FMD) between generated and target embedding distributions.
\item \textbf{Control quality}: style linearity error and monotonicity violation from $g(\hat{m}_\alpha)$ trajectories.
\item \textbf{Extrapolation}: ExtraErr against held-out over-reaction motion $m_{s_2}$ at high alpha.
\end{itemize}

\subsubsection{Controllability metrics.}
For each sample, we evaluate $\alpha \in \{0,0.5,1.0,1.5,2.0\}$ and compute style-strength trajectory $g(\hat{m}_\alpha)$. We report:
\begin{equation}
\text{Mono}=\frac{1}{K-1}\sum_{k=1}^{K-1}\mathbb{I}\!\left[g(\hat{m}_{\alpha_{k+1}})\ge g(\hat{m}_{\alpha_k})\right],
\end{equation}
\begin{equation}
\text{MonoViol}=\frac{1}{K-1}\sum_{k=1}^{K-1}\max\!\left(0, g(\hat{m}_{\alpha_k})-g(\hat{m}_{\alpha_{k+1}})\right),
\end{equation}
\begin{equation}
\text{LinErr}=\frac{1}{K}\sum_{k=1}^{K}\left|g(\hat{m}_{\alpha_k})-\alpha_k g(m_s)\right|.
\end{equation}
In tables, we report monotonicity violation (\texttt{MonoViol}), i.e., averaged positive decreases in $g$ across consecutive alphas.

\subsubsection{Extrapolation metric.}
At high intensity, we compare $\hat{m}_{2.0}$ with held-out $m_{s_2}$ in the same embedding space:
\begin{equation}
\text{ExtraErr}=\left\|\phi(\hat{m}_{2.0})-\phi(m_{s_2})\right\|_2,
\end{equation}
where $\phi(\cdot)$ is the frozen TMR style feature extractor.

\subsubsection{Standard transfer metrics.}
Following prior style-transfer evaluations~\cite{song2024mcmldm}, we report FMD, CRA, and SRA for compatibility with existing baselines.

\subsection{Main Quantitative Results}
Table~\ref{tab:main_results} shows the global comparison. Our method gives the best control-related behavior: highest SRA (0.295), lowest MonoViol (0.05), lowest LinErr (0.280), and lowest ExtraErr (1.109). MCM-LDM remains strong on FMD and CRA, indicating a realism/content advantage under the current setting, while our method better preserves intensity ordering and style-strength controllability.

Table~\ref{tab:per_dataset} shows the per-dataset behavior on the three main style datasets. On PerMo, our SRA and monotonicity are stronger than MCM-LDM (SRA: 0.140 vs.\ 0.080; MonoViol: 0.03 vs.\ 0.22). On Bandai-Namco and Xia, our method keeps lower LinErr; MonoViol is lower on Bandai-Namco and slightly higher on Xia, indicating that controllability gains vary with dataset style geometry.

\begin{table}[t]
  \centering
  \caption{\textbf{Main quantitative results.} FMD/CRA/SRA/MonoViol/LinErr are averaged over PerMo/Bandai-Namco/Xia; ExtraErr is evaluated on the over-reaction set.}
  \label{tab:main_results}
  \begin{tabular}{lcccccc}
    \toprule
    Method & FMD $\downarrow$ & CRA $\uparrow$ & SRA $\uparrow$ & MonoViol $\downarrow$ & LinErr $\downarrow$ & ExtraErr $\downarrow$ \\
    \midrule
    DeepMotionEditing~\cite{Aberman2020} & 0.588 & 0.290 & 0.060 & 0.25 & 0.326 & 1.196 \\
    MCM-LDM~\cite{song2024mcmldm} & \textbf{0.381} & \textbf{0.808} & 0.265 & 0.11 & 0.358 &  1.174 \\
    Ours & 0.457 & 0.606 & \textbf{0.295} & \textbf{0.05} & \textbf{0.280} & \textbf{1.109} \\
    \bottomrule
  \end{tabular}
\end{table}

\begin{table}[t]
  \centering
  \caption{\textbf{Per-dataset results.} Each cell is \textit{Ours/MCM-LDM}.}
  \label{tab:per_dataset}
  \begin{tabular}{lccccc}
    \toprule
    Dataset  & FMD $\downarrow$ & CRA $\uparrow$ & SRA $\uparrow$ & MonoViol $\downarrow$ & LinErr $\downarrow$ \\
    \midrule
    PerMo & 0.531 / 0.474 & 0.700 / 0.880 & 0.140 / 0.080 & 0.03 / 0.22 & 0.421 / 0.464 \\
    Bandai-Namco & 0.143 / 0.237 & 0.714 / 1.000 & 0.714 / 0.429 & 0.01 / 0.06 & 0.259 / 0.368 \\
    Xia & 0.279 / 0.346 & 0.108 / 0.351 & 0.405 / 0.216 & 0.07 / 0.05 & 0.261 / 0.407 \\
    \bottomrule
  \end{tabular}
\end{table}

\subsection{Extrapolation to 2$\times$ Style Targets}
We evaluate out-of-range behavior on our over-reaction captures by comparing generated $\hat{m}_{2.0}$ against real $m_{s_2}$ motions. This isolates whether the model genuinely learns a style-intensity axis beyond trained endpoints.
On the over-reaction dataset, ExtraErr is 1.109 (ours), 1.196 (DeepMotionEditing), and 1.174 (MCM-LDM), indicating that ours is closest to the real stronger-style targets.

\subsection{Ablation Studies}
Table~\ref{tab:ablation} shows the effect of direction mode and loss terms on PerMo. The instance direction gives the best FMD, SRA, and LinErr in this setting, while the hybrid direction remains competitive on CRA and MonoViol. Removing either regularizer weakens controllability: without $\mathcal{L}_{\text{lin}}$, FMD/CRA/SRA degrade; without $\mathcal{L}_{\text{mono}}$, CRA/SRA also drop.

\begin{table}[t]
  \centering
  \caption{\textbf{Ablation study (PerMo).} Effect of direction mode and loss terms.}
  \label{tab:ablation}
  \resizebox{\linewidth}{!}{%
  \begin{tabular}{lcccccccc}
    \toprule
    Variant & $\mathcal{L}_{\text{lin}}$ & $\mathcal{L}_{\text{mono}}$ & Direction & FMD $\downarrow$ & CRA $\uparrow$ & SRA $\uparrow$ & MonoViol $\downarrow$ & LinErr $\downarrow$ \\
    \midrule
    Ours & Yes & Yes & instance & \textbf{0.531} & \textbf{0.70} & \textbf{0.14} & 0.03 & \textbf{0.421} \\
    Dir=proto & Yes & Yes & proto & 0.584 & 0.60 & 0.06 & 0.04 & 0.513 \\
    Dir=hybrid ($\beta=0.3$) & Yes & Yes & hybrid & 0.588 & \textbf{0.70} & 0.12 & 0.03 & 0.509 \\
    w/o $\mathcal{L}_{\text{lin}}$ & No & Yes & instance & 0.605 & 0.54 & 0.06 & 0.03 & 0.422 \\
    w/o $\mathcal{L}_{\text{mono}}$ & Yes & No & instance & 0.545 & 0.56 & 0.04 & 0.03 & 0.434 \\
    \bottomrule
  \end{tabular}%
  }
\end{table}

\subsection{Qualitative Results}
\begin{figure}[t]
  \centering
  \includegraphics[width=\linewidth]{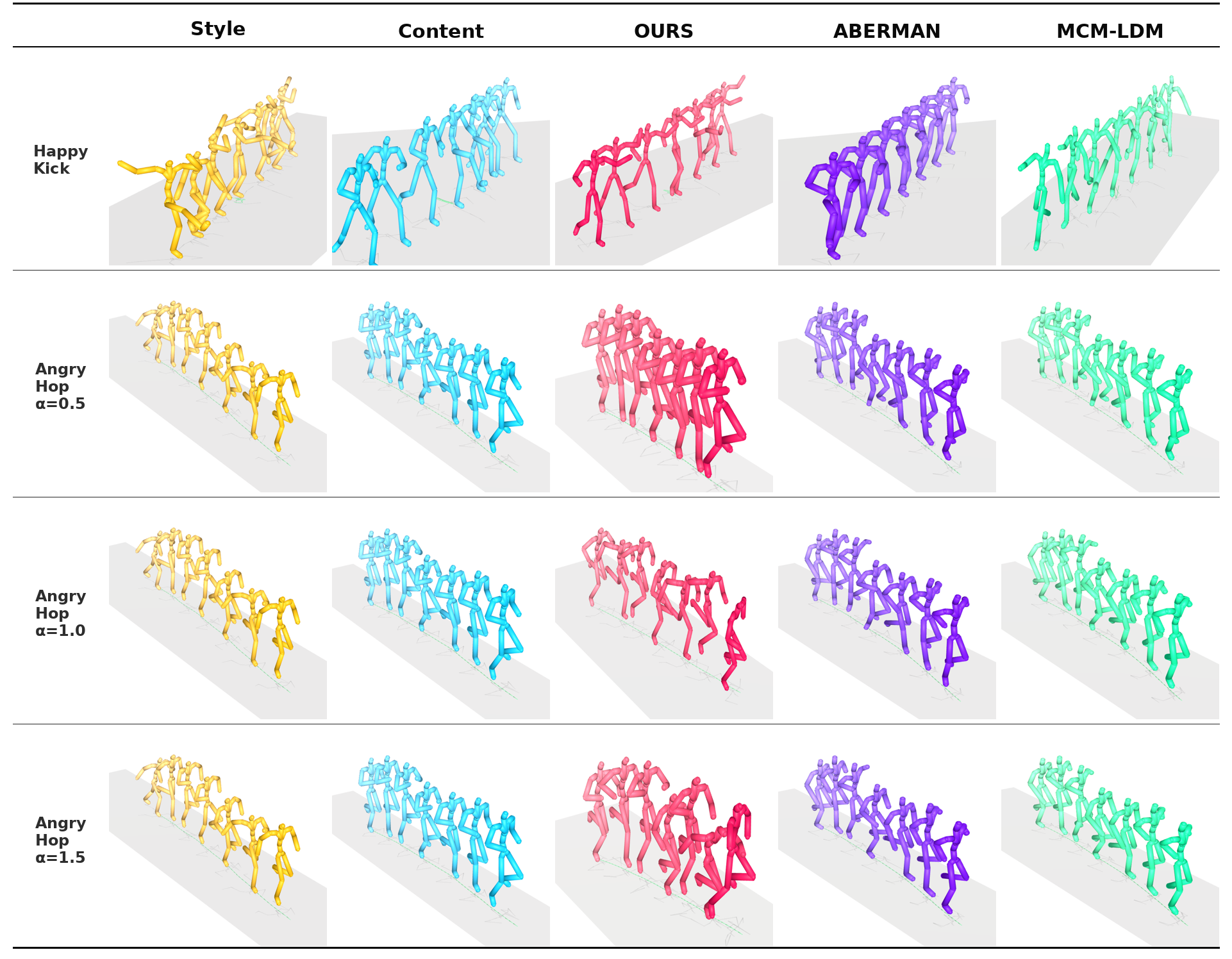}
  \caption{\textbf{Qualitative comparison.} Columns show style reference, content reference, and generated results from our method and baselines under a unified camera and rendering configuration.}
  \label{fig:qual_grid}
\end{figure}

Figure~\ref{fig:qual_grid} provides side-by-side visualizations at matched camera/view settings. We observe that our method produces clearer intensity separation while preserving motion identity, especially in difficult cases with different motion amplitudes (e.g., kick vs.\ hop). This visual trend is consistent with the higher SRA and lower LinErr/MonoViol in Tables~\ref{tab:main_results} and~\ref{tab:per_dataset}.

\begin{figure}[t]
  \centering
  \includegraphics[width=\linewidth]{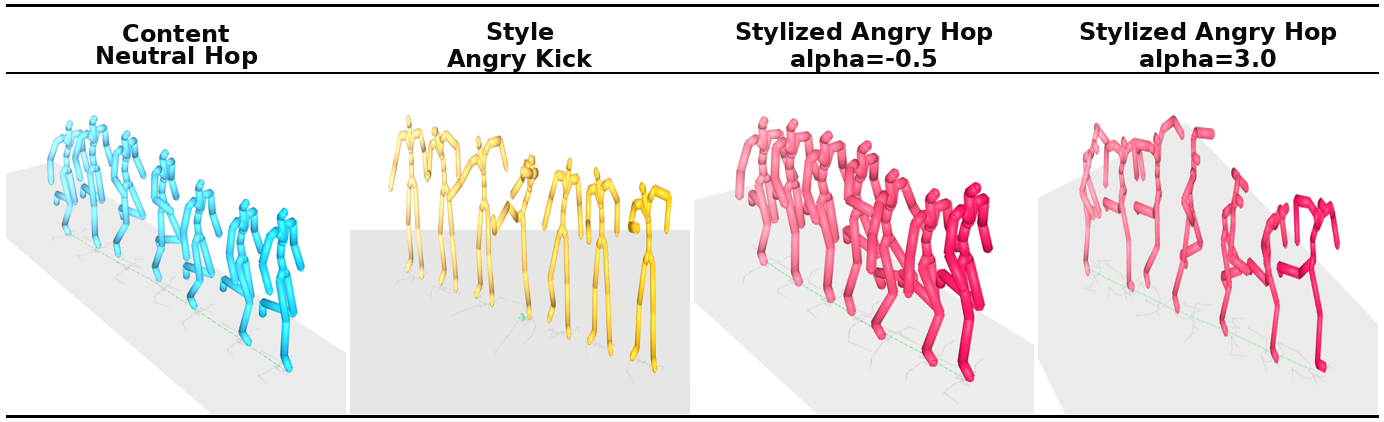}
  \caption{\textbf{Cross-content stylization example.} Although our quantitative benchmark uses endpoint-paired motions to keep evaluation clean, the learned style direction can also be applied qualitatively across different content motions. Here, a neutral hop is edited using an angry kick reference; negative and large positive $\alpha$ values modulate the output along this reference direction while retaining the hopping content.}
  \label{fig:cross_content}
\end{figure}

Figure~\ref{fig:cross_content} illustrates a broader use case beyond the strict paired evaluation setting. We do not claim fully general text- or arbitrary-reference style transfer as the main benchmark objective; rather, this example shows that the reference-based slider can produce meaningful cross-content behavior, supporting its practical use as an animation editing control.

\subsection{User Study}
We conducted a Likert-scale user study with 11 university participants and no monetary compensation. The study focused on PerMo and compared our method with MCM-LDM using 12 questionnaire sections (6 motion cases $\times$ 2 methods). Each section showed one triplet generated from the same content--style pair at $\alpha\in\{0.5,1.0,1.5\}$. The three videos were anonymized as A/B/C. 
The A/B/C-to-$\alpha$ assignment was randomized per section, and participants were not told which method generated each triplet. Participants answered seven questions: stylization strength for A/B/C, naturalness for A/B/C, and one overall question on whether the style-intensity differences were distinguishable. All ratings used a 7-point Likert scale, where larger values indicate stronger style, higher naturalness, or clearer distinguishability. For analysis, A/B/C were remapped to true low/mid/high $\alpha$. Detailed user study protocols can be found in the supplementary material.

Table~\ref{tab:user_study} summarizes results on matched sections ($n=6$). Our method shows consistently higher stylization scores at all intensity levels (low: $3.92$ vs.\ $3.56$, mid: $4.61$ vs.\ $3.59$, high: $5.18$ vs.\ $3.46$) and much higher distinguishability ($4.92$ vs.\ $2.36$). Naturalness is comparable between methods (ours: $4.64/4.83/4.53$, MCM-LDM: $4.74/4.71/4.71$ for low/mid/high). 
For perceived monotonic control, we count a motion case as a pass when the mean participant ratings satisfy style\_low < style\_mid < style\_high after remapping the shuffled A/B/C videos to their true alpha order. As the result, ours achieves a higher pass rate ($0.50$ vs.\ $0.00$).

Paired sign tests show the same trend but limited statistical significance due to small sample size: $\Delta$style\_high $=+1.73$ ($p=0.2188$), $\Delta$distinguishability $=+2.56$ ($p=0.0625$), and monotonic-pass improvement $=+0.50$ ($p=0.2500$).

\begin{table}[t]
  \centering
  \small
  \caption{\textbf{User study summary (6 matched motion cases, 11 participants).} Scores are 1--7 Likert means of low/mid/high (L/M/H) intensity per section.}
  \label{tab:user_study}
  \begin{tabular}{lccc}
    \toprule
    Method & Distinguish $\uparrow$ & Stylization (L/M/H) $\uparrow$ & Naturalness (L/M/H) $\uparrow$ \\
    \midrule
    Ours & \textbf{4.92} & \textbf{3.92 / 4.61 / 5.18} & 4.64 / \textbf{4.83} / 4.53\\
    MCM-LDM & 2.36 & 3.56 / 3.59 / 3.46 & \textbf{4.74} / 4.71 / \textbf{4.71}\\
    \bottomrule
  \end{tabular}
\end{table}

\section{Discussion}
\label{sec:discussion}

Our results should be interpreted with the task definition in mind: the goal is not only to produce one stylized sample, but to provide a predictable editing axis for style strength. For this setting, SRA, LinErr, MonoViol, and ExtraErr are direct measures of whether the slider follows the requested style direction and intensity ordering. CRA and FMD remain important supporting metrics for content preservation and motion realism, but they do not by themselves measure whether the user can reliably ask for ``less'', ``more'', or extrapolated style.

This distinction is important in practical animation workflows. Artists frequently adjust style strength iteratively, and a method that generates plausible individual samples but responds inconsistently to the control value is difficult to use as an editing interface. Our results therefore support the central hypothesis that style intensity should be treated as a relative control coordinate rather than a fixed global scale. The cross-content result in Figure~\ref{fig:cross_content} further suggests that the same reference-based control signal can be useful beyond exactly paired motions, while our quantitative evaluation keeps the endpoint-paired setting to ensure clean, reproducible measurement.

The over-reaction benchmark is useful beyond headline scores. It separates interpolation quality from true out-of-range behavior by evaluating against real stronger-style targets $m_{s_2}$. In this setting, methods can have similar in-range style recognition but different extrapolation error and monotonic behavior, indicating that endpoint supervision alone is insufficient unless the conditioning geometry is well constrained. The lower ExtraErr of our method suggests that learning a consistent style direction not only improves interpolation but also helps maintain a meaningful trajectory when extrapolating beyond the observed endpoints.

\subsubsection{Limitations.}
First, controllability depends on the quality of endpoint pairs and the style embedding calibration used by both training regularizers and evaluation. Noisy or weakly matched pairs reduce direction quality. Second, large-$\alpha$ generation may still produce motion artifacts for rare styles or sparse actors. A possible remedy is to add kinematic or physical constraints (e.g., foot-contact consistency and motion smoothness penalties) and curriculum training that gradually increases the sampled $\alpha_{\max}$. Third, our quantitative benchmark focuses on reference-based endpoint control; extending the slider to unseen-style or text-guided control is an important direction for future work.

\section{Conclusion}
\label{sec:conclusion}
We presented Motion Style Slider, an endpoint-supervised framework for continuous style-intensity control in motion diffusion. The core design combines latent style-direction conditioning with a scalar control parameter and explicit regularization for linearity and monotonicity, enabling controllable interpolation and extrapolation without intermediate-intensity supervision.

Across the PerMo, Bandai-Namco, and Xia motion dataset, our method improves control-oriented behavior (style fidelity and slider consistency) while remaining competitive on realism/content metrics. We further introduced an over-reaction extrapolation benchmark that evaluates against real stronger-style motions, and used it to measure out-of-range performance directly rather than only in-range transfer quality.

Overall, these results support the view that practical motion stylization needs a reliable relative control axis rather than a single fixed style strength. Future work includes stronger biomechanical constraints for high-$\alpha$ stability, improved cross-dataset style alignment, and larger user studies.

\section*{Acknowledgements}
We thank Osaka Cygames' Motion Capture Studio for capturing the motion dataset and for their support throughout the project.

%
%
\bibliographystyle{splncs04}
\bibliography{main}
\end{document}